\documentclass[letterpaper, 10 pt, conference]{ieeeconf}  

\IEEEoverridecommandlockouts                              

\usepackage{kotex}
\usepackage{amsmath}
\usepackage{mathtools}
\usepackage{svg}
\usepackage{xcolor}
\usepackage{cite}
\usepackage{cleveref}
\usepackage{multirow}
\usepackage{booktabs}
\usepackage{float}

\title{\LARGE \bf
SIMS: Surgeon-Intention-driven Motion Scaling for Efficient and Precise Teleoperation
}

\author{Anonymous Anonymous$^1\dagger$ 
\thanks{1\dagger Anonymous}
}

\author{Jeonghyeon Yoon$^{1*}$, Sanghyeok Park$^{2*}$, Hyojae Park$^{1}$, Cholin Kim$^{1}$, Micheal Yip$^{3}$, and Minho Hwang$^{1\dagger}$
\thanks{* Equal Contribution, ${\dagger}$ Corresponding author}
\thanks{$^1$Department of Robotics and Mechatronics Engineering, DGIST, Daegu, 42988, Republic of Korea
        {\tt\footnotesize \{yjh1434, hyojae, cholin, psh120, minho\}@dgist.ac.kr}}%
\thanks{$^2$School of Undergraduate Studies, DGIST, Daegu, 42988, Republic of Korea
        {\tt\footnotesize sh.park@dgist.ac.kr}}%
\thanks{$^3$Department of Electrical and Computer Engineering, University of California San Diego, La Jolla, CA 92093 USA.{yip}@ucsd.edu}%
}

\begin{document}

\crefname{figure}{Fig.}{Figs.}
\maketitle
\thispagestyle{empty}
\pagestyle{empty}

\begin{abstract}

\textbf{Telerobotic surgery often relies on a fixed motion scaling factor (MSF) to map the surgeon’s hand motions to robotic instruments, but this introduces a trade-off between precision and efficiency: small MSF enables delicate manipulation but slows large movements, while large MSF accelerates transfer at the cost of accuracy. We propose a \textbf{Surgeon-Intention driven Motion Scaling (SIMS)} system, which dynamically adjusts MSF in real time based solely on kinematic cues. SIMS extracts linear speed, tool motion alignment, and dual-arm coordination features to classify motion intent via fuzzy C-means clustering and applies confidence-based updates independently for both arms. In a user study (\(n=10\), three surgical training tasks) conducted on the da Vinci Research Kit, SIMS significantly reduced collisions (mean reduction of 83\%), lowered mental and physical workload, and maintained task completion efficiency compared to fixed MSF. These findings demonstrate that SIMS is a practical and lightweight approach for safer, more efficient, and user-adaptive telesurgical control.
}
\end{abstract}

\section{INTRODUCTION}

When the operator’s intuitive workspace differs from the restricted workspace where the actual task is performed, telerobotic scalability has been employed as an effective solution. Such discrepancies are not limited to simple positional misalignments but often manifest as significant volumetric differences between the two workspaces. Telerobotics systems are typically based on a leader–follower architecture, in which the motion generated by the operator through the leader robot is transmitted to the remote follower robot after undergoing a scaling transformation. This enables the operator to maintain intuitive control while effectively overcoming workspace mismatches.

The medical field represents a prominent example where telerobotics has successfully addressed workspace mismatches. The da Vinci Surgical System converts centimeter-scale hand movements of the surgeon into millimeter-scale precise motions of surgical instruments, enabling minimally invasive procedures within the narrow abdominal cavity \cite{b1,b2}. Likewise, microsurgical robots such as the Symani Surgical System achieve micrometer-level motion scaling, allowing surgeons to perform demanding procedures like vascular or lymphatic anastomosis \cite{b3,b4}. In in-vitro fertilization (IVF), cell injection robots transform coarse user inputs into micrometer-level injection motions, thereby enhancing consistency and improving success rates of the procedure \cite{b5,b6}.

\setlength{\textfloatsep}{10.0pt plus 2.0pt minus 2.0pt}
\begin{figure}[!t]
    \centering
    \includegraphics[width=\linewidth]{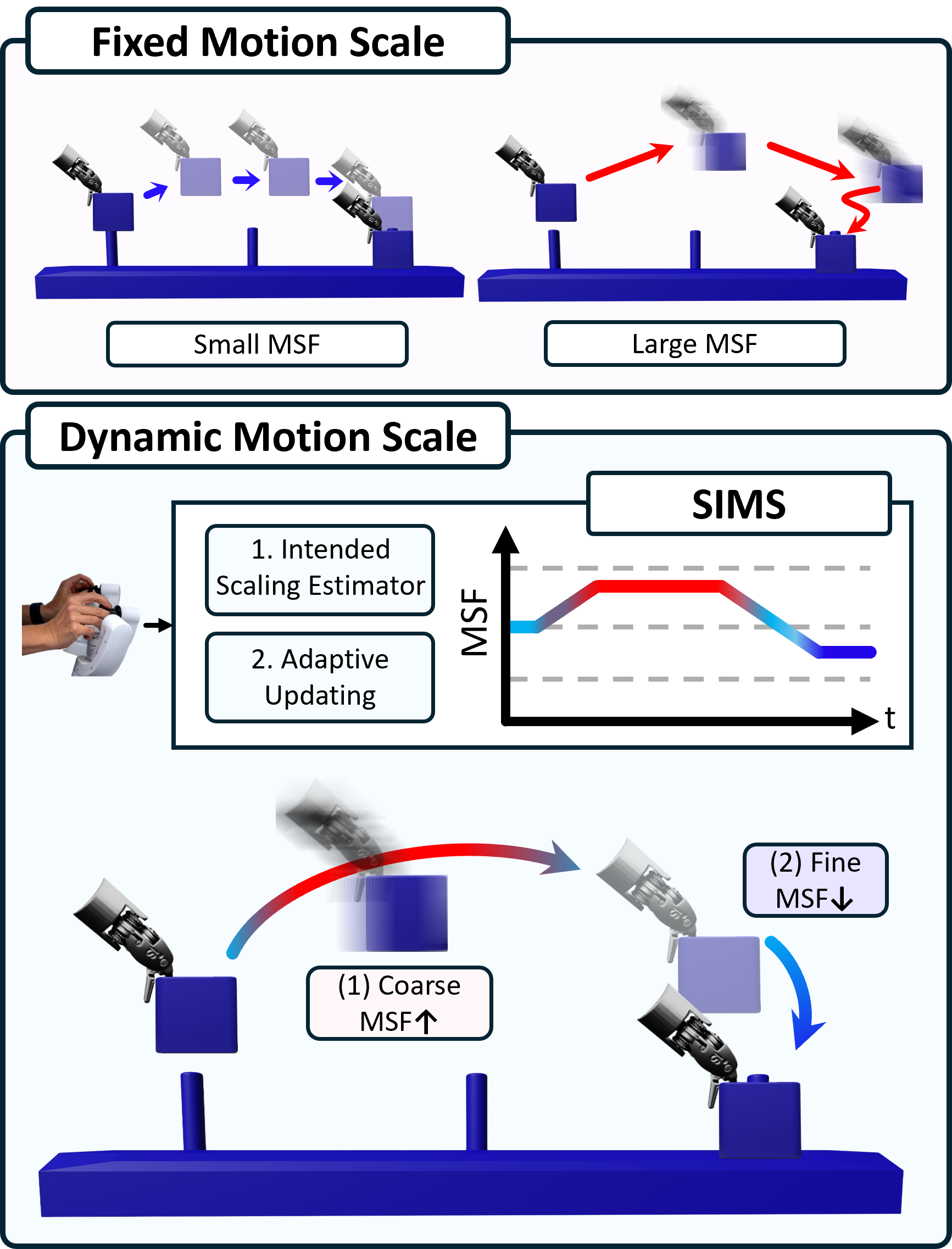}
    \caption{\textbf{Overview of SIMS.} Conventional teleoperation relies on a fixed MSF, where a small MSF ensures precision but slows down large movements, and a large MSF allows fast movements but sacrifices accuracy (top). The proposed SIMS dynamically adjusts the MSF based on the surgeon’s intent: when coarse motion is detected, the MSF increases to enable efficient movement, and when fine motion is required, it decreases to ensure precision (bottom).}
    \label{fig:intro}
\end{figure}

Most existing telesurgical robotic systems employ a fixed motion scaling factor (MSF), yet this approach exhibits several limitations in diverse clinical contexts. Surgical tasks require frequent alternation between coarse motions for rapid instrument positioning and fine motions for delicate incisions or suturing. A fixed MSF cannot adequately satisfy both demands. A smaller MSF (e.g., 0.3) faithfully reflects subtle hand movements and ensures precision but slows instrument movement, potentially prolonging the surgery. Conversely, a larger MSF (e.g., 3.0) facilitates fast instrument transfer but compromises precision during delicate tissue manipulation, posing safety risks. Consequently, surgeons are forced to compensate for inappropriate scaling through either excessively fine or exaggerated hand motions, which increases both cognitive and physical workload, leading to fatigue and reduced concentration during long procedures.

To overcome the trade-off between precision and efficiency imposed by fixed MSF, a new approach that adaptively adjusts the MSF according to the situation is required. In this study, we draw inspiration from the concept of human motor skills, which are generally categorized into gross motor skills and fine motor skills \cite{b7}. The former corresponds to large, fast, and coarse motions, whereas the latter corresponds to small, precise, and fine motions. This distinction directly relates to the motion scaling problem in telesurgical systems. For example, in minimally invasive surgery (MIS) or microsurgery, stages requiring rapid instrument repositioning benefit from a larger MSF, while stages involving delicate tissue manipulation require a smaller MSF.

This paper presents an Surgeon-Intention driven Motion Scaling (SIMS) system for telesurgical robots. An overview of the proposed SIMS system is illustrated in \cref{fig:intro}. The system relies solely on kinematic data to infer the surgeon’s intended motion scale, enabling computationally efficient and real-time adaptation. A soft-clustering–based framework dynamically adjusts the MSF according to user characteristics, ensuring both precision and responsiveness. Experiments on the da Vinci Research Kit (dVRK) demonstrate superior performance over conventional fixed scaling with improved task execution. Main contributions in this study include:

\begin{itemize}
    \item Compact kinematic descriptors relevant to robotic surgery are formulated for motion scale classification, designed for real-time operation without relying on visual inputs.
    \item A soft-clustering–based real-time framework enables dynamic and user-specific motion scaling, allowing the MSF to adapt continuously to the surgeon’s intent.
    \item User study on the dVRK shows that SIMS significantly reduces collisions, lowers perceived workload, and maintains task efficiency, demonstrating improved safety and usability over fixed MSF.
\end{itemize}

\section{Related Works}

MSF is particularly crucial, as it directly contributes to surgical efficiency and safety by regulating the ratio between the surgeon’s hand motions and the corresponding movements of the robotic instruments. According to Parsa et al., the absence of motion scaling led to significantly higher mental demand compared to conditions with motion scaling, as indicated by a mean difference of 8.889 (95\% CI [1.282, 16.496])\cite{b19}. Similarly, Cassilly et al. found that a fixed 1:1 motion scaling ratio can lead to a higher number of errors and negatively impact operator performance\cite{b20}.

To address the limitations of a fixed MSF, various studies have explored dynamically adjusting the MSF based on the context of the surgical task. One common strategy involves using spatial information. Heredia-Pérez et al. proposed a region-based system where the workspace is predefined with a triangular mesh, and the MSF is adjusted based on the end-effector's proximity to the nearest mesh region, each assigned a specific scaling value \cite{b21}. Other approaches rely on kinematic features. Richter et al. introduced a velocity-based method where the MSF is adjusted proportionally to the speed of the master input device \cite{b22}. Building on this, Lim et al. expanded their research by proposing a method to determine the optimal MSF based on real-time communication latency and user-specific characteristics \cite{b23}. In a different line of work, leader–follower teleoperation is modeled as a human-in-the-loop control system to determine the MSF through filter design \cite{b24}.

Existing approaches are often limited by heavy reliance on workspace or single kinematic feature, restricting their ability to provide continuous and personalized scaling. By contrast, the proposed SIMS system infers coarse–fine motion intent solely from kinematic data and adapts the MSF in real time through soft clustering, thereby reducing computational cost while seamlessly bridging coarse and fine motions to enhance surgical efficiency.


\begin{figure*}[t!]
    \centering
    \includegraphics[width=\linewidth]{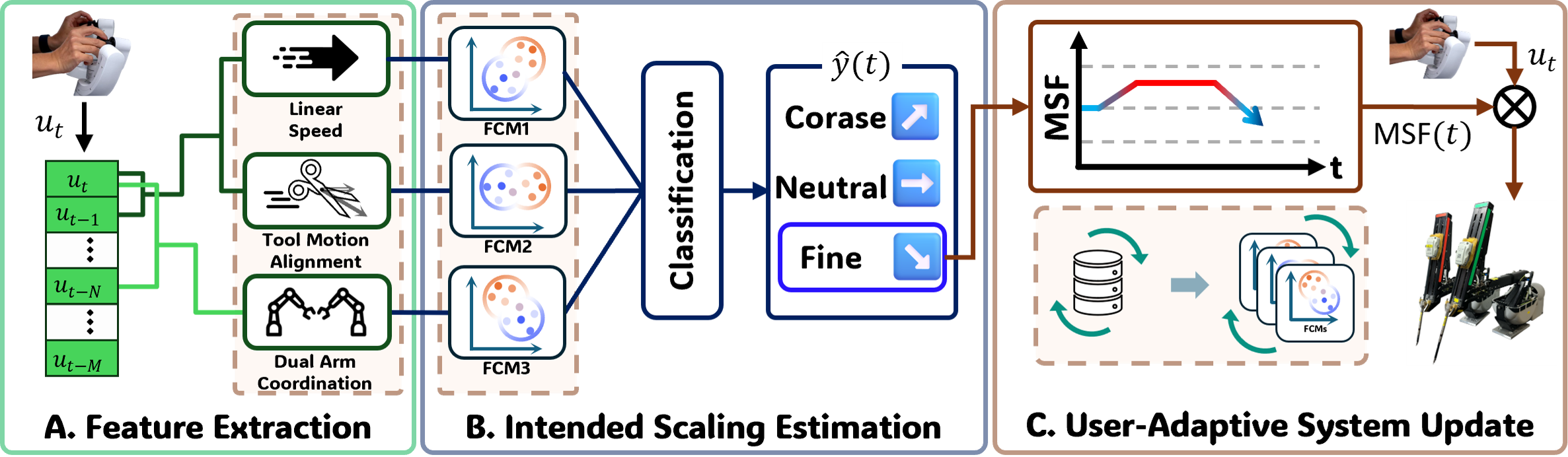}
    \caption{\textbf{System Diagram of SIMS.} SIMS infers the surgeon’s intended motion scale (coarse, neutral, fine) from hand-controller inputs and dynamically adjusts the MSF. 
\textbf{A. Feature Extraction:} Real-time pose commands ($u_t$) generate trajectories, from which linear speed, tool motion alignment, and dual-arm coordination are extracted. 
\textbf{B. Intended Scaling Estimation:} Pre-trained FCM models output membership values for each motion class, which are fused to estimate the intended scale at 60 Hz during teleoperation. 
\textbf{C. User-Adaptive Update:} The inferred scale determines the ramp function’s sign and magnitude for MSF updates: fine decreases MSF, coarse increases MSF, and neutral maintains it. Features are buffered, and when sequence length and variance criteria are met, FCM models are updated with recent data to reflect user adaptation while preserving diversity.}

    \label{fig:intent}
\end{figure*}

\section{System Architecture}

In this study, SIMS was implemented and evaluated on dVRK, a teleoperated surgical robot. In this setup, the surgeon manipulates hand controllers of the master tool manipulators (MTMs) to operate the patient-side manipulators (PSMs). The surgeon’s hand motions are scaled through an MSF before being executed by the PSMs, thereby enabling precise instrument manipulation within the constrained surgical workspace.

Building on this framework, SIMS dynamically adjusts the MSF according to the surgeon’s intent and motion characteristics. The overall system configuration of SIMS is shown in \cref{fig:intent}. The system is organized around three main components: (1) surgical task–specific kinematic feature extraction, (2) a soft-clustering–based intended scaling estimator that infers the desired motion scale and applies it smoothly to the robot, and (3) an adaptive update mechanism that enables the classifier to evolve as more data become available. Details of each component are provided in the following subsections.

\subsection{Feature Extraction}

In this work, only kinematic data were used to define the features $\{f_i(t), i \in \{1,2,3\}\}$ required for motion scale classification. This approach ensures low computational cost for real-time implementation while reflecting the characteristics of surgical robot manipulation. The defined features are as follows:

\begin{itemize}
    \item \textbf{Linear Speed ($f_1$)}: The tool center point (TCP) linear velocity is the most direct indicator for distinguishing coarse from fine motion. Higher velocities indicate coarse movements, whereas lower velocities represent fine manipulation.  
    \item \textbf{Tool Motion Alignment ($f_2$)}: The alignment between the TCP velocity vector and the tool direction reflects motion characteristics. Precise tool alignment, such as during fine needle grasping, corresponds to fine motion, while misalignment during tool transport corresponds to coarse motion.  
    \item \textbf{Dual-Arm Coordination ($f_3$)}: Surgical tasks such as needle handover, suturing, and tissue dissection require precise coordination between both arms. Stronger coordination results in smaller differences in velocity profiles. A time-window–based formulation is employed to suppress the effect of local tremor or momentary imbalance.  
\end{itemize}

\begin{table}[H]
\caption{Extracted Features for Motion Scale Recognition}
\label{MSF_feature}
\centering
\begin{tabular}{@{}l l c c@{}} 
\toprule
\multicolumn{3}{c}{\textbf{Features, $f$}} & \textbf{Equation, $f_i(t)$} \\
\midrule
Linear Speed           
    & \rule{0pt}{4ex}\( f_{1}(t) = \|\boldsymbol{v}(t)\| \)\\
Tool Motion Alignment       
    & \rule{0pt}{4ex}\( f_{2}(t) = 1 - \left|\frac{\boldsymbol{v}(t)\cdot \boldsymbol{d}(t)}{\|\boldsymbol{v}(t)\|\|\boldsymbol{d}(t)\|}\right| \)\\
Dual-arm Coordination    
    & \rule{0pt}{4ex}\( f_{3}(t) = \frac{\sum_{k=t-N}^{t} |m_L(k) - m_R(k)|}{\sum_{k=t-N}^{t} (m_L(k) + m_R(k)) + \epsilon} \)\\
\bottomrule
\multicolumn{4}{l}{\begin{tabular}[c]{@{}l@{}}
\footnotesize
* $\boldsymbol{v}(t)$: Linear velocity of the TCP. \\
* $\boldsymbol{d}(t)$: Unit vector representing the tool direction. \\
* $m_L(k), m_R(k)$: Linear/rotational speed of the left and right arms. \\
* $\epsilon$: A small positive constant for numerical stability. \\
* $N$: Size of the time window
\end{tabular}}
\end{tabular}
\end{table}

The mathematical formulations of these features, extracted from the end-effector trajectory of the surgical robot, are summarized in Table~\ref{MSF_feature}. All features are normalized to the range $[0,1]$, where smaller values indicate fine motion and larger values indicate coarse motion. These features are then used as inputs to the soft-clustering framework described in the next subsection. While each feature individually provides useful cues for classification, relying on a single feature can lead to misclassification due to sensitivity to specific task conditions or surgeon-specific manipulation styles. Therefore, a multi-feature approach combining linear velocity, tool motion alignment, and dual-arm coordination is adopted to robustly infer the surgeon’s intended motion scale.

\subsection{Intended Scaling Estimation}
The purpose of this module is to infer the surgeon’s intended motion scale. To this end, feature trajectories extracted from the past manipulation data are used to train a \textit{Fuzzy C-Means} clustering (FCM) model. Unlike hard clustering methods, FCM assigns \textit{membership values} that indicate the degree to which each trajectory belongs to multiple clusters simultaneously. This property captures the inherent ambiguity in human intent and provides a more flexible representation than forcing a single motion class assignment.  

Given a feature vector $\mathbf{f}_i(t)$, the membership value $m_{ij} \in [0,1]$ for motion class $j \in \{\text{fine}, \text{neutral}, \text{coarse}\}$ is computed as:  
\begin{equation}
u_{ij} = \frac{1}{\sum_{k=1}^{C}\left(\frac{\|\mathbf{f}_i - c_j\|}{\|\mathbf{f}_i - c_k\|}\right)^{\tfrac{2}{m-1}}}  
\end{equation}

where $c_j$ is the cluster center, $C$ is the number of clusters, and $m$ is the fuzzification parameter. Further details of FCM can be found in~\cite{b25}. During training, three clusters (fine, neutral, and coarse motion) are established.  

During real-time inference, trajectories are input into three feature-specific FCM models, producing class-wise membership values. The membership values for each motion class are then averaged across the three features (late fusion), and the class with the maximum weighted average membership is selected as the final intended motion class:  
\begin{equation}
\hat{y} = \arg\max_j \left( \frac{1}{3}\sum_{i=1}^3 w_{i}u_{ij} \right)  
\end{equation}

where \(w_i\) is a user-defined weight for the feature class \(i\), enabling adjustment of each motion class's influence in the final decision.

\subsection{User-Adaptive System Update}

For the system to operate adaptively, it is crucial that the FCM-based inference continuously reflects the surgeon’s manipulation style and adaptation to the environment. To this end, two design strategies are introduced: \textit{data selection} and \textit{confidence-based MSF update}.  

\textbf{Data Selection.}  
As the surgeon repeatedly performs tasks, motion patterns evolve as they adapt to the robot and its dynamics. Simply using all available data uniformly for training limits the ability to reflect the surgeon’s most recent manipulation style. To overcome this limitation, a data selection strategy is employed: the most recent M trajectories are prioritized, while a fraction of past trajectories is retained to ensure diversity. This strategy allows (i) rapid adaptation to the surgeon’s latest behavior and (ii) robustness against overfitting to temporary noise or task-specific fluctuations. The length of the training sequence ($M$) can be tuned experimentally. 

Training sequences are distinguished based on the degree of dispersion across three features. The Quartile Coefficient of Dispersion (QCD) was employed as a measure of relative variability, defined as:

\begin{equation}
QCD = \frac{Q_3 - Q_1}{Q_3 + Q_1}
\end{equation}

where $Q_1$ and $Q_3$ denote the first and third quartiles, respectively. As $QCD$ captures the relative spread of the data distribution, it serves as an effective indicator of feature diversity. Once a trajectory comprising at least minimum data points has been accumulated, $QCD$ values are computed for each feature. If any of these values exceed the predefined threshold, the dataset is deemed to exhibit sufficient diversity. At this point, a new training sequence is established from the accumulated trajectories, and trajectory accumulation is re-initiated. The length of minimum data and threshold of $QCD$ are determined empirically.

Consequently, the proposed data selection mechanism captures the surgeon’s adaptation process while maintaining stable generalization performance.  

\textbf{Confidence-based MSF Update.}  
The inferred intended motion scale $\hat{y}(t)$ is applied to update the MSF through a confidence-based adjustment, where confidence corresponds to the normalized membership value of the winning cluster in the FCM, thereby reflecting the reliability of the classification.
\begin{equation}
\text{MSF}(t) = 
\begin{cases}
\text{MSF}(t-1) + \Delta, & \hat{y}(t) = \text{coarse}, \\[4pt]
\text{MSF}(t-1) - \Delta, & \hat{y}(t) = \text{fine}, \\[4pt]
\text{MSF}(t-1), & \hat{y}(t) = \text{neutral},
\end{cases}
\end{equation}
\begin{equation}
\Delta = W_{\min} + \frac{\sum_{i=1}^3w_iu_{i\hat{y}}}{\sum_{i=1}^3w_i}\big(W_{\max} - W_{\min}\big)
\end{equation}

where $W_{\min}$ and $W_{\max}$ specify the minimum and maximum update magnitudes, and $\hat{y}$ denotes the final intended motion class. The fraction in the $\Delta$ definition corresponds to the normalized membership value of the winning cluster, ranging from 0 (least confident) to 1 (most confident). By adapting $\Delta$ in proportion to this confidence, the MSF update becomes more aggressive when the classification is reliable, while remaining conservative under uncertainty. This mechanism allows the system to achieve both rapid adaptability to the surgeon’s intention and robustness against transient noise.

In the proposed framework, all processes are executed independently for the left and right arms. The final MSF is then determined through an OR operation that selects the larger scaling factor between the two arms. For example, if the left arm is classified as coarse while the right arm is neutral or fine, the overall MSF is set to coarse and applied equally to both arms. Likewise, if one arm is neutral and the other is fine, the system selects neutral as the final MSF. This mechanism ensures that the overall scaling factor always reflects the larger motion requirement between the two arms, thereby maintaining consistency and stability in bimanual teleoperation.
     
\begin{figure}[t]
    \centering
    \includegraphics[width=\linewidth]{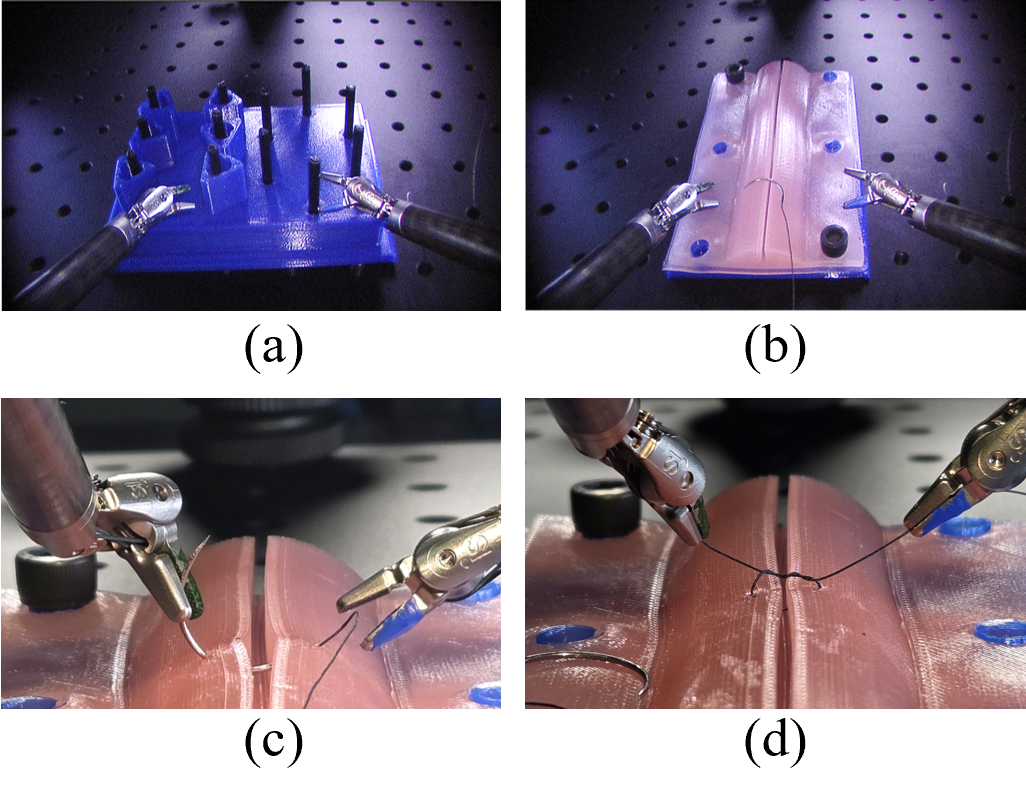}
    \caption{\textbf{Evaluation Surgical Tasks for SIMS and Fixed MSF Comparison.}
    Experimental tasks: (a) peg transfer, (b) phantom tissue setup, (c) surgical stitching, and (d) surgical knot tying.}
    \label{fig:experiment_tasks}
\end{figure}

\begin{table*}[t]
\centering
\caption{Performance Metrics under Different Motion Scaling Conditions. }
\resizebox{\textwidth}{!}{%
\begin{tabular}{ll|c|cccc}
\toprule
\textbf{Task} & \textbf{Condition} & \textbf{Collision Count} (Safety) ↓ & \textbf{Clutch Count} ↓ & \textbf{Task Completion Time [s]} ↓ & \textbf{Path Length [m]} ↓ & \textbf{Motion Smoothness [\(m^2/s^5\)]} ↓ \\
\midrule
\multirow{3}{*}{Peg Transfer} 
 & Small MSF & 0.42 $\pm$ 0.49 & 4.15 $\pm$ 3.83 & 101.50 $\pm$ 31.67 & 1.70 $\pm$ 0.39 & \textbf{4.88 $\pm$ 1.40} \\
 & Large MSF & 1.50 $\pm$ 0.96 & \textbf{0.30 $\pm$ 0.56} & \textbf{64.89 $\pm$ 13.60} & 1.94 $\pm$ 0.39 & 5.79 $\pm$ 1.62 \\
 & SIMS      & \textbf{0.17 $\pm$ 0.37} & 2.10 $\pm$ 1.81 & 80.55 $\pm$ 20.69 & \textbf{1.94 $\pm$ 0.40} & 6.23 $\pm$ 2.02 \\
\midrule
\multirow{3}{*}{Knot Tying} 
 & Small MSF & 0.75 $\pm$ 0.72 & 1.00 $\pm$ 1.99 & 78.41 $\pm$ 30.69 & 1.44 $\pm$ 0.53 & 6.31 $\pm$ 2.69 \\
 & Large MSF & 1.00 $\pm$ 0.91 & \textbf{0.05 $\pm$ 0.22} & \textbf{52.42 $\pm$ 23.99} & 1.51 $\pm$ 0.61 & \textbf{5.68 $\pm$ 1.66} \\
 & SIMS      & \textbf{0.17 $\pm$ 0.37} & 0.68 $\pm$ 1.03 & 53.80 $\pm$ 25.38 & \textbf{1.15 $\pm$ 0.41} & 5.86 $\pm$ 2.37 \\
\midrule
\multirow{3}{*}{Stitching} 
 & Small MSF & 0.25 $\pm$ 0.43 & 0.93 $\pm$ 1.66 & 61.67 $\pm$ 28.72 & \textbf{0.98 $\pm$ 0.34} & \textbf{6.18 $\pm$ 2.01} \\
 & Large MSF & 0.75 $\pm$ 0.83 & \textbf{0.00 $\pm$ 0.00} & \textbf{39.58 $\pm$ 16.93} & 1.01 $\pm$ 0.30 & 6.24 $\pm$ 2.12 \\
 & SIMS      & \textbf{0.17 $\pm$ 0.37} & 0.80 $\pm$ 1.52 & 53.44 $\pm$ 40.99 & 1.01 $\pm$ 0.53 & 6.91 $\pm$ 2.68 \\
\bottomrule
\end{tabular}}
\vspace{0mm}
\footnotesize
\begin{flushleft}
* Values represent  $\text{mean} \pm\text{standard deviation}$. \\
* A total of 10 participants performed each MSF condition (Small MSF, Large MSF, SIMS) across three tasks, with four repetitions per task. \\
* Bold numbers indicate the best values.
\end{flushleft}
\label{tab:msf_results}
\end{table*}


\section{Experiment and Result}

To validate the effectiveness of the proposed SIMS method, we conducted a user study ($n=10$) using a physical surgical robotic platform dVRK in a realistic teleoperation environment. All participants had prior experience with teleoperation interfaces, ensuring familiarity with robot-assisted manipulation tasks. Before the experiment, each participant was given approximately three hours of practice time to become proficient in controlling the dVRK \cite{dvrk} system via teleoperation. The dVRK was used to compare SIMS with a fixed MSF baseline across multiple standardized surgical training tasks selected from the Fundamentals of Laparoscopic Surgery (FLS) curriculum \cite{FLS}, as shown in \cref{fig:experiment_tasks}. This section first describes the experimental setup, followed by details of the task design and procedure, and concludes with the experimental results and analysis.

\subsection{Experimental Setup}

Experiments were performed on the dVRK in teleoperation mode, comprising a pair of MTMs for surgeon input, two PSMs for instrument control, and an Endoscopic Camera Manipulator (ECM) equipped with a stereo endoscope. Stereo endoscopic images were captured at a resolution of 1920×1080 pixels at 60 fps and synchronized with robot kinematic data, including joint positions, Cartesian end-effector poses, and motion scaling states. All data were recorded at 60 Hz to ensure precise alignment between visual feedback and motion information. A total of $n=10$ participants with prior teleoperation experience but without formal surgical training performed the experiments in a benchtop training environment using standardized laparoscopic training models. Both the proposed SIMS algorithm and the fixed MSF baseline were implemented within the same teleoperation control framework to ensure consistent comparisons.

\subsection{Tasks and Metrics}

To evaluate the effectiveness of the proposed SIMS system, three standardized tasks from the FLS curriculum—peg transfer, stitching, and knot tying—were selected. These tasks were chosen to capture a wide range of teleoperation demands: peg transfer emphasizes tool–tool coordination and workspace coverage, stitching tests fine motion precision and repeatability, and knot tying requires complex multi-step manipulation and dexterity.

Before testing SIMS, participants additionally performed two repetitions of the peg transfer task under a nominal MSF to collect training data for the FCM models used to initialize SIMS. After this initialization step, participants repeated the same tasks under three MSF conditions: (i) fixed small scale (\(MSF = 0.2\)), (ii) fixed large scale (\(MSF = 0.4\)), and (iii) SIMS (adaptive scaling between 0.2 and 0.4). These scaling values were chosen based on the workspace ratio between MTMs and PSMs of the dVRK to represent clinically meaningful ranges. The order of tasks and scaling conditions was randomized to minimize learning effects.

Each task was repeated four times per condition, with one trial defined as follows:
\begin{itemize}
    \item \textbf{Peg transfer:} Move all blocks from one side of the pegboard to the opposite side (one full board transfer) per trial.
    \item \textbf{Stitching:} Pass a needle through tissue once and completely pull the attached thread through per trial.
    \item \textbf{Knot tying:} Tie one surgeon's knot per trial.
\end{itemize}

Performance was evaluated using six quantitative metrics and one subjective measure:
\begin{itemize}
    \item \textbf{Collision count:} Instrument–environment or instrument–instrument contact events, normalized per trial.
    \item \textbf{Clutch count:} Number of clutch activations and total clutch duration per trial.
    \item \textbf{Task completion time:} Total time to complete each trial.
    \item \textbf{Path length:} Total end-effector trajectory length per trial.
    \item \textbf{Motion smoothness:} Mean squared jerk of the end-effector trajectory, 
    \(\text{Smoothness} = \frac{1}{T} \int_0^T \|\dddot{\mathbf{x}}(t)\|^2 dt\),
    where jerk \(\dddot{\mathbf{x}}(t)\) was estimated via finite differences of 60~Hz position data.
    \item \textbf{NASA-TLX:} Post-experiment survey of perceived workload\cite{nasa-tlx}.
\end{itemize}

\begin{figure}[t]
    \centering
    \includegraphics[width=\linewidth]{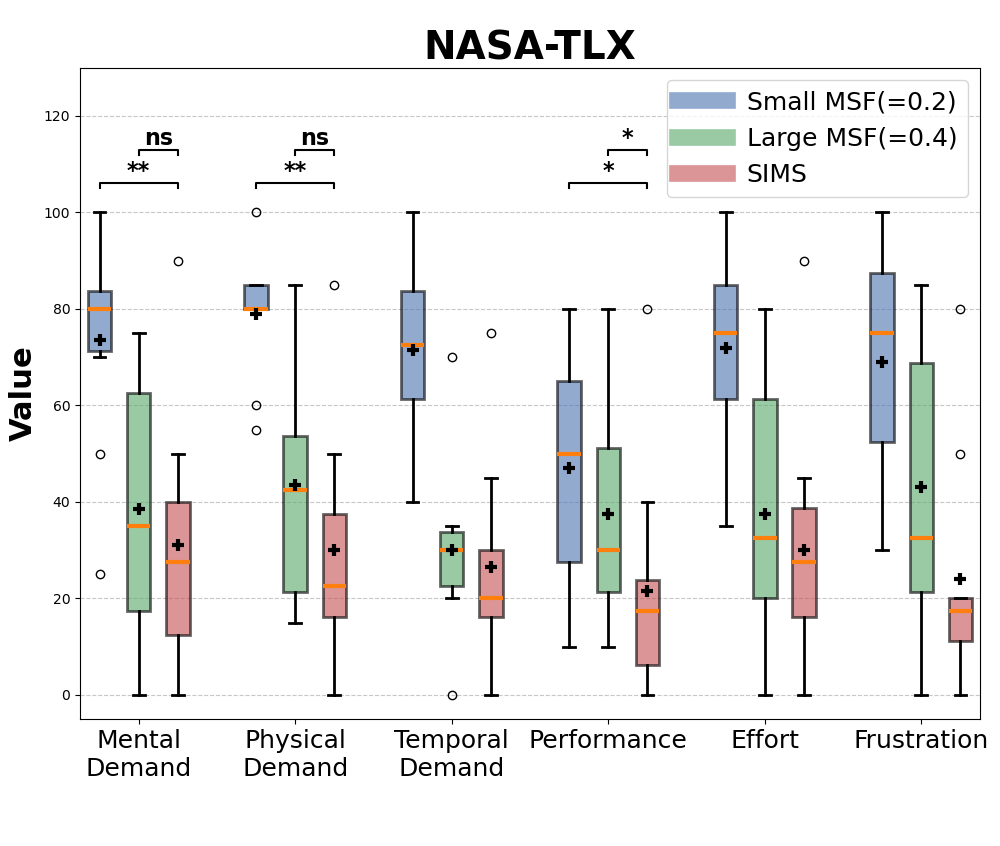}
    \caption{\textbf{NASA-TLX workload survey}
    The results across six subscales (mental, physical, temporal demand, performance, effort, frustration). SIMS shows consistently lower perceived workload compared to fixed MSF settings. * indicates $p < 0.05$, ** indicates $p < 0.01$, and ns denotes not significant.}
    \label{fig:workload}
\end{figure}

\begin{figure}[t]
    \centering
    \includegraphics[width=\linewidth]{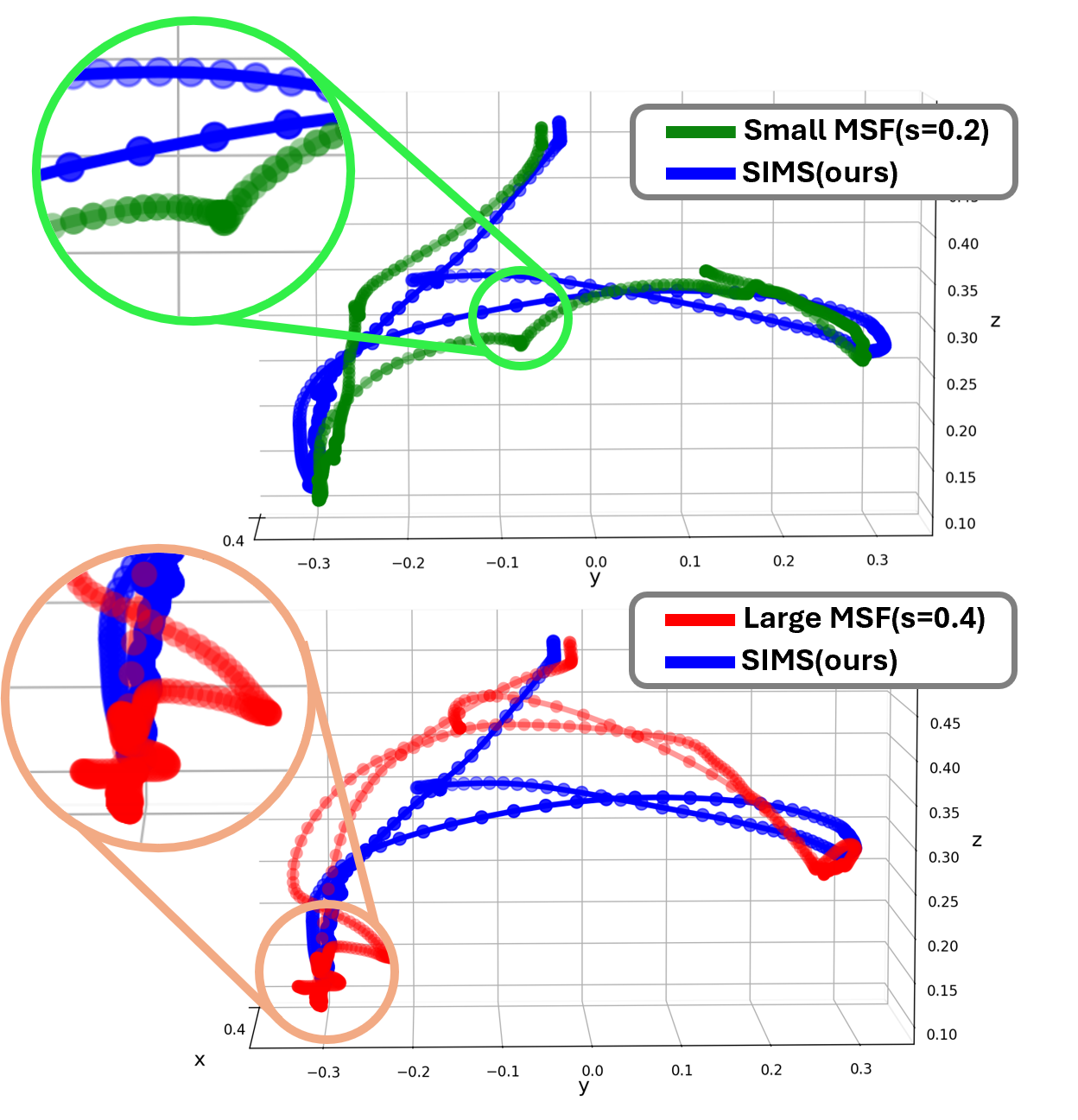}
    \caption{\textbf{End-effector Trajectory}
    Representative end-effector trajectories during the peg transfer task under three configurations. Fixed small scale (green, top) required frequent clutching, leading to fragmented and repetitive motion. Fixed large scale (red, bottom) caused over-amplification and jittery movements, increasing collision risk. SIMS (blue) maintained smooth, continuous trajectories with reduced clutching and improved control stability.}
    \label{fig:trajectory}
\end{figure}

\subsection{Results and Analysis}

Table~\ref{tab:msf_results} summarizes the quantitative metrics, \cref{fig:workload} illustrates the subjective workload (NASA-TLX), and \cref{fig:trajectory} presents representative end-effector trajectories under different MSF conditions. All metrics are reported as \(\text{mean} \pm \text{standard deviation}\) over all trials. Each participant performed all three tasks (Peg Transfer, Knot Tying, Stitching) under each condition (Small MSF, Large MSF, SIMS) with four repetitions per task.

Compared to the small MSF condition, SIMS consistently outperformed in all aspects—efficiency, safety, and workload—demonstrating its ability to reduce clutching and task time while minimizing collisions and perceived effort. When compared to the large MSF condition, SIMS provided clear advantages in safety and workload, while achieving comparable efficiency without a substantial performance drop. Overall, these results show that SIMS offers a well-balanced trade-off, combining the safety and precision of small scaling with much of the efficiency of large scaling. The following sections present a more detailed breakdown of these results, highlighting how SIMS effectively balances safety, efficiency, and user experience.

\textbf{1) Safety–Efficiency Trade-off:}  
The table highlights the inherent trade-off between speed and safety when using fixed MSF. Small MSF consistently minimized overshoot and ensured fine control but required frequent clutching and significantly increased task completion times (TCT). Large MSF achieved faster task execution and fewer clutch activations but at the cost of higher collision counts and less stable trajectories.  
SIMS consistently produced the lowest collision counts across all tasks, matching or closely approaching the efficiency of large MSF in terms of TCT and path length. By balancing these metrics, SIMS demonstrates that adaptive scaling can achieve near-optimal speed while significantly improving safety, reducing the need for manual clutching.

\textbf{2) User Workload and Motion Quality:}  
NASA-TLX results (\cref{fig:workload}) confirm that SIMS lowered perceived workload across all subscales. 
Compared to the small MSF condition, these reductions were statistically significant in both Mental and Physical Demand ($p<0.01$), as well as in Performance ($p<0.05$). 
When compared to the large MSF condition, SIMS still showed lower demand scores in magnitude, though the differences did not reach statistical significance, while Performance was rated significantly better than with large MSF ($p<0.05$).
Although SIMS did not achieve the absolute best values in every objective efficiency metric, participants consistently reported a more comfortable and controllable teleoperation experience, suggesting improved user satisfaction. 
Furthermore, compared to large MSF, SIMS preserved much of its speed advantage while reducing cognitive and physical strain, demonstrating that adaptive scaling enhances usability even without maximizing every performance measure. 
Trajectory analysis (\cref{fig:trajectory}) further shows that SIMS maintained smooth, continuous end-effector motion, avoiding abrupt corrections and oscillations seen with large MSF, thereby contributing to more stable and intuitive teleoperation.

\textbf{3) Task-specific Insights:}  
In simpler transport-focused tasks like peg transfer, SIMS demonstrated clear benefits, reducing collisions by over 80\% while retaining comparable speed to large MSF. For complex tasks such as knot tying and stitching, SIMS achieved collision rates much lower than either fixed scaling setting, though its timing performance remained close to large MSF rather than outperforming it. These results suggest that while expert users may prefer a fixed, conservative scaling for extremely fine manipulations, adaptive scaling offers a safer and more versatile baseline for general teleoperation.  

SIMS consistently achieves a balanced performance profile, excelling in safety metrics without compromising efficiency. This makes it a promising, lightweight solution for surgical teleoperation, where both speed and safety are critical. Future work will explore hybrid strategies that combine SIMS with context recognition to selectively switch between adaptive and fixed scaling for specific surgical subtasks.

\section{CONCLUSIONS}

This work presented the Surgeon-Intention driven Motion Scaling (SIMS) system, a real-time teleoperation framework that dynamically adjusts the MSF using only kinematic cues. SIMS leverages compact features—linear speed, tool motion alignment, and dual-arm coordination—combined with a FCM framework and confidence-based MSF updates to provide adaptive, user-specific scaling without visual feedback. Experiments with 10 participants performing three surgical training tasks on the da Vinci Research Kit demonstrated that SIMS effectively balances safety and efficiency, reducing collisions by over 80\% while maintaining task completion times comparable to large fixed scaling.

Although SIMS offers a lightweight and low-latency solution, this study also highlights the need for further research to extend its adaptability. Current results show that SIMS excels at providing a safe baseline for general teleoperation, but additional intelligence is required to optimize scaling for complex, high-skill subtasks such as knot tying and precise suturing. Future work will focus on integrating context-aware models, such as visual-language or transformer-based approaches, to recognize task phases and dynamically adjust scaling strategies. Combining these capabilities with SIMS’s low-overhead clustering framework could enable a hybrid control system that seamlessly switches between adaptive and fixed scaling modes, personalizes responses to individual surgeons, and further improves safety and efficiency in real surgical workflows.

\addtolength{\textheight}{-12cm}   





\bibliographystyle{IEEEtran}
\bibliography{ref}

\end{document}